# Combining RGB and Points to Predict Grasping Region for Robotic Bin-Picking

Quanquan Shao[a], Jie Hu
Shanghai Jiao Tong University
Shanghai, China
e-mail: [a]sjtudq@qq.com

*Abstract*—This paper focuses on a robotic picking tasks in cluttered scenario. Because of the diversity of objects and clutter by placing, it is much difficult to recognize and estimate their pose before grasping. Here, we use U-net, a special Convolution Neural Networks (CNN), to combine RGB images and depth information to predict picking region without recognition and pose estimation. The efficiency of diverse visual input of the network were compared, including RGB, RGB-D and RGB-Points. And we found the RGB-Points input could get a precision of 95.74%.

*Keywords-component; RGB-Points, suction grasp, nueral networks, bin-picking*

## I. INTRODUCTION

Robotic picking objects in clutter are widely researched in recently years. Relative technologies have a wide range of application demands in material transportation, waste sorting and logistics automation. However, it remains a big challenge because of the variety of scenarios, the diversity of objects, clutter by placing and complicated background. Vision based robotic manipulation have been successfully used in various applications in manufacturing industry and logistics industry. Normally, vision based robotic grasping methods mainly include object detection, object segmentation, pose estimation and the selection of grasping point[1]. 2D features are used to estimate the pose of target objects[2][3][4]. With the development of 3D vision technology, 3D visions are also used in robotic grasp in cluttered scene to detect the objects and estimate poses[5][6]. Matching 3D models for object recognition and pose estimation is difficult with various target objects and even impossible for unknown objects because of the less of 3D models beforehand[7]. Self-occlusion and disordered placing also weaken the performance of RGB-D based part 3D matching in robotic picking tasks. Traditional technical routes are weak to deal with grasp tasks in cluttered scenario, especially with unknown objects.

Compared with model-based technologies, data-driven methods have a great success in many vision tasks. Deep Convolution Neural Networks (CNN) got a great performance in image classification tasks [8]. Deep Neural Networks (DNN) are also used in robotic grasp tasks. Vision based system with DNN could choose the grasp point in cluttered scenario directly without 3D models or pose estimation[9][10]. These methods treat the grasp point detection as classification problems or regress the coordinates of grasp point directly after training. DNN based grasp point detection could not only grasp known objects but also grasp unknown objects with a high performance of generalization. There are mainly three types of hands involved in robotic grasp tasks, namely, dexterous hand, three or two fingers grippers and suction gripper. Most of grasping point detection are based on parallel-jaw grasping configuration. Other gripper configurations are not studied deeply, especially suction gripper. In this paper, we focus on suction grasps which is also widely applied in robotic grasp and object manipulation. A region prediction approach was proposed to detect suction point in cluttered scenario with known or unknown target objects. Visual information got by a RGB-D camera was inputted into a U-net structure convolution neural network and the output is a probability map which means the successful probability of each pixel as a suck point. After some smoothing and other post process with this probability map, we could choose a best suck point in bin-picking like cluttered environment. We expanded the input of region detection networks with RGB-D and RGB-Points The result are shown below and we could find the RGB-Points have the best performance.

The paper is organized as follows: we first review related works in section 2, and then elaborately present our region suction point detection method with a deep neural network in U-net framework in section 3. Finally, we show our experiments and conclusion in section 4 and section 5.

## II. RELATED WORK

Object grasp has a long history in robotics research and it is still an active research topic until now. Early studies mainly research the pose estimation and the grasp plan with force-closure and form closure. Rahardja used 2D image feature to recognize the objects and estimate their poses in plane coordinates in bin-picking situations[2]. RANSAC[11] is widely used in pose estimation with image feature or point clouds for object picking in cluttered scene[6][12][13][14]. ICP[15] are also often used in object recognition and pose estimation with point clouds[16][17].

Most of researches are based on force-closure and form closure in grasp plan field in past decades[18]. The stability of grasped objects in force unclosed situation is also studied[19]. Force-close with the effect of friction is also explored[20]. More details of grasping configuration could be found in[21]. Nevertheless, All these methods are model-based and need the template to match. They have weak performance in complicated background and lack the ability to deal with unknown object. Learning based methods are also used in synthesizing grasp configuration. These methods could predict the grasp point directly without pose estimation or 3D model of target objects[22]. With the great success of deep neural network in vision tasks, most of learning based grasp detection methods use neural networks to process vision information in recent research[9][10][23].

Rectangles were used stand for grasp configuration and image features were learned automatically to detect robotic grasps with RGB-D information in situation with objects placed separately[23]. Some variant was proposed to improve the speed of this learning-based method. Pinto used the same grasp configuration and studied grasp detection with self-supervision type, which was low efficient and needed 700 hours[9]. Levine combined reinforcement learning with vision robotic grasp with millions grasp trials. As a result, the robot could grasp objects placed in a bin without hand-eye calibration[24]. Konstantinos improved this method with GAN and domain adaption using simulation to speed up the learning speed[25]. However, learning from scratch is low efficient and time-consuming. Levine's method and its variant are all only using the RGB images without depth information.

Pas used point clouds got by depth sensor to generate lots of grasp candidates with some projection process and evaluated these candidates with a convolution neural network[10]. These methods described above are all in situation with parallel-jaw or multi-finger grippers. Mahler introduced a Grasp Quality Convolution Neural Network (GQ-CNN) for estimating the quality of suction with point clouds, which used 1500 3D object models to train the network[26]. Zeng applied fully convolutional networks to suction grasp detection and took the first place in Amazon Robotics Challenge[27]. However, it needed reprojection before inputting prediction network and used the height map to predict grasp region. Inspired by above research, we directly use the perception information of RGB-D camera to input prediction network in suction gripper configuration. At the same time, we compare the performance of RGB images, RGB image with depth maps and RGB-Points. And we could find that combining RGB images with point clouds have the best performance.

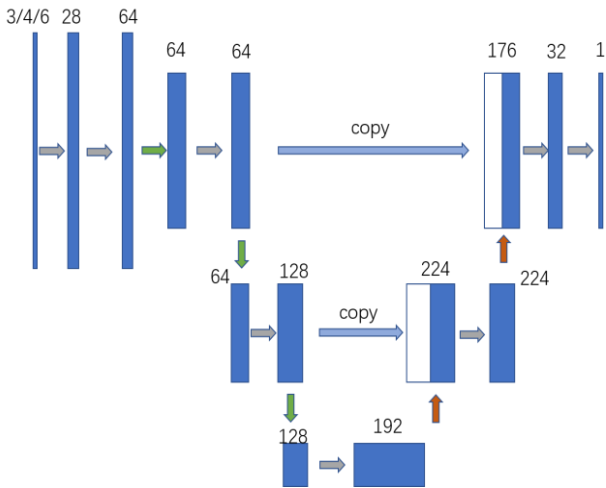

Figure 1: The Framework of U-net

III. MATHODOLOGY

The perception network has a U-net framework which combines down-pooling operation and up-sample operation shown as Figure 1. Left-to-right grey arrow means a layer of convolution neural network. Top-to-bottom green arrow means a layer of max-pooling down-sampling features and bottom-to-top brown arrow means a layer of deconvolution operation which could expand the size of feature map. The input of U-net is a RGB-D camera perception information, which includes RGB image, depth image or point clouds. The output of this neural network is the successful suction region probability map. Each pixel value of this probability map denotes predicted successful possibility while sucking vertically at the 3D point corresponding to this pixel point.

The U-net framework had been successfully used in image segmentation and region detection with images in different situations. In this paper, we use U-net to predict the suction grasp region map with visual information directly in robotic picking system. The framework is mainly composed by convolutional layers which could learn multifold scale features expression by itself avoiding the complex artificial feature design. The detail architecture of this region prediction neural network is shown in Table 1. The size of input images is 128×128 while the size of output is 64×64. Some operations of batch normalization are executed after each convolutional layer, which are not expressed in the table. To train the U-net, we collect training data with an Intel Realsense camera, which gets color image and depth map with a 30Hz frequency. The points could also be obtained after a simple process with projection matrix of the sensor. The relation between point clouds and depth map is as follows:

$$\begin{bmatrix} x \\ y \\ z \end{bmatrix} = z\Sigma^{-1} \begin{bmatrix} u \\ v \\ 1 \end{bmatrix} \quad (1)$$

where $\Sigma$ is projection matrix of the RGB sensor, $[x, y, z]^T$ is coordinate of point in RGB local frame, z is the value in the depth map and $[u, v, 1]^T$ stands for generalized coordinate of each point in depth map. What we need to point out is that the depth map has been registered into the RGB frame with official RealSense SDK. And then we labeled these images as mask maps manually (shown in Figure 2).

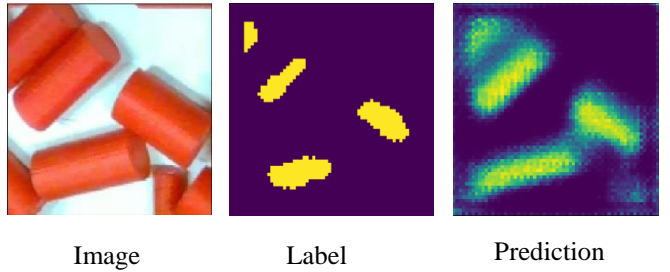

Image     Label     Prediction

Figure 2: Labeled images and prediction

Graspable region was labeled as one while other area was labeled as zero. We totally collected 950 labeled images and 800 images were used for training this grasp region prediction network. And the rest of these images were used for testing and validation purpose.

Table 1: The Structure of U-net

| Number | Layers | Parameters |
|---|---|---|
| 1. | Input | Images, Depth, Points |
| 2. | convolution1 | Kernel: 28; Size: 11×11 |
| 3. | convolution2 | Kernel: 64; Size: 7×7 |
| 4. | max-pooling1 | Stride: 2; Size: 2×2 |
| 5. | convolution3 | Kernel: 64; Size: 5×5 |
| 6. | max-pooling2 | Stride: 2; Size: 2×2 |
| 7. | convolution4 | Kernel: 128; Size: 3×3 |
| 8. | max-pooling3 | Stride: 2; Size: 2×2 |
| 9. | convolution5 | Kernel: 192; Size: 3×3 |
| 10. | deconvolution1 | Kernel: 192; Size: 2 × 2 |

| 11. | convolution5 | Kernel: 224; Size: 3×3 |
| 12. | deconvolution2 | Kernel: 224; Size: 2 × 2 |
| 13. | convolution6 | Kernel: 176; Size: 3×3 |
| 14. | convolution7 | Kernel: 32; Size: 3×3 |
| 15. | convolution8 | Kernel: 1; Size: 5×5 |
| 16. | output | Sigmoid function |

As the labeled data was limited, we also used some data augmentation tricks such as flip and rotation to expand the labeled data in training phase. And then U-net could be trained with these labeled images. A weight cross-entropy loss function with weight decay was chosen in this situation:

$$\text{Loss} = \alpha y(-\log(\tilde{y})) - (1-y)\log(1-\tilde{y}) + \beta w^2 \quad (2)$$

where α is weight of positive in cross entropy, $\beta$ is weight of regularization, $w$ is parameter of the network while y, $\tilde{y}$ is label value and prediction value respectively.

The next step was choosing the suction point with the predict map. As shown before, the map meant the probability of graspable region. The graspable region was partly disconnected. We used a gaussian filter to smooth the result and focus on the center of the graspable region. The gaussian filter was followed:

$$Gaus = \frac{1}{16}\begin{bmatrix} 1 & 2 & 1 \\ 2 & 4 & 2 \\ 1 & 2 & 1 \end{bmatrix} \quad (3)$$

After that normalized operation was executed. At last we chose the max value point of the processed map as the grasp point.

## IV. EXPERIMENT AND RESULTS

The network was structured with Tensorflow1.0, a machine learning system published by Google. And the hardware is a notebook with a 2.6GHz Intel Core i7-6700HQ CPU and a NVDIA GTX 965 GPU. The learning rate is 0.001 with a decay 0.8. The weight of positive pixel, namely α in the loss function, is 5, 4, 2 respectively with different inputs. And the weight of regularization ($\beta$) is $10^{-4}$. The sensor is Intel Realsense SR300 RGB-D camera. The resolution of color image is $1920 \times 1080$ while the resolution of infrared camera is $640 \times 480$. The detect range of depth is 0.2m to 1.5m with a precision 0.125 millimeter.

We trained the U-net with different inputs and compared the performance of these inputs. There was color image (RGB), color image with depth (RGB-D) and color image with point clouds (RGB-Points). The point clouds were calculated with the depth image and the projection matrix of the visual sensor. As the point clouds were special inputs, we first chose a boundary of these points and normalized coordinates of these points to 0 ~ 1 respectively. The points beyond this boundary were set to zeros. Because of the precision of the depth sensor, there are many points having null depth value. We set x-coordinate, y-coordinate and z-coordinates of these points are all zeros. These points were all in range from 0 to 1 and could be processed like images.

Traditionally, PR-curves was used to evaluate the performance of the saliency detection and segmentation However, recall rate was not important. We chose the precision rate as the evaluation index. After the normalization of prediction map, different threshold value was chosen to evaluate the performance.

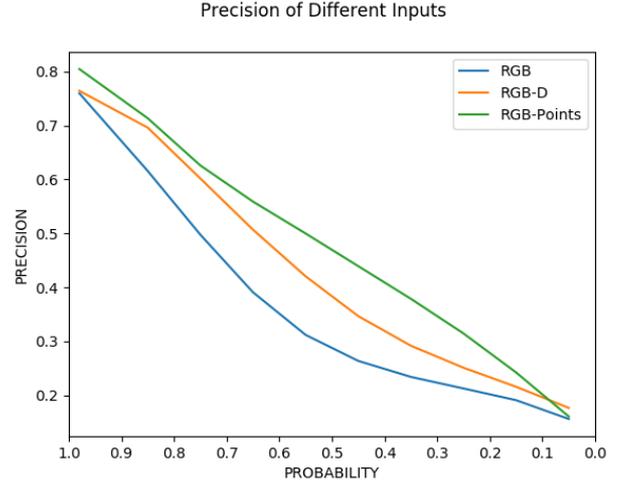

Figure 3: The Results of Different Inputs

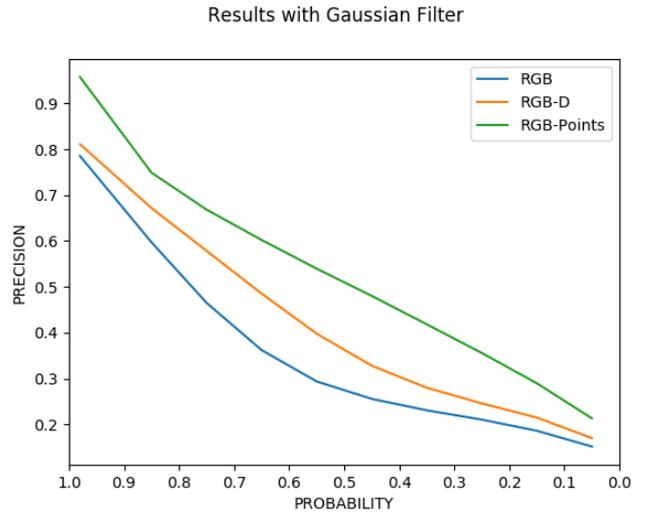

Figure 4: The Results with Gaussian Filter

$$Prec = \frac{num(pred >= threshold)}{num(label=1)} \quad (4)$$

The results of these experiments were shown in Table 2 and Figure 3. We could find that color images with point clouds have the best performance compared with RGB and RGB-D inputs. The depth dimension could also improve the performance of the prediction of grasp region in cluttered scene.

Table 2: The Results of Different Inputs

| Threshold | RGB | RGB-D | RGB-Points |
|---|---|---|---|
| 0.98 | 0.7597 | 0.7643 | 0.8046 |
| 0.85 | 0.6157 | 0.6960 | 0.7134 |

We also compared the performance of gaussian filter post process. The results could be found in Figure 3. The result with gaussian smooth was shown in Figure 4 and Table 3. Compared Table 2 with Table 3, we could find that post process of prediction map improved the performance of the prediction in all three different inputs. With the threshold 0.98, the precision value of color image with point clouds was 95.74%.

We also constructed a suction grasping platform with a real robot in ROS environment. The framework of grasping platform was shown in Figure 5.

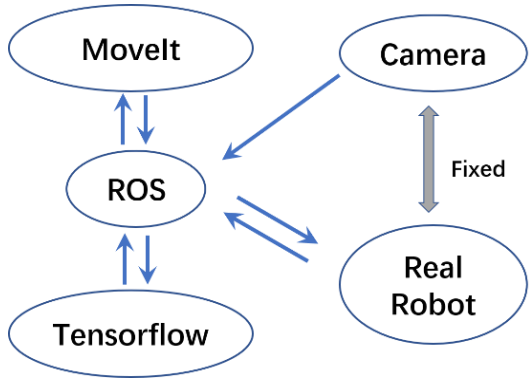

Figure 5: The Framework of Suction Platform

The camera got images and depth maps and they were sent to ROS with a ROS message. Grasping prediction node operated these images to get the suction point with tensorflow toolkits. After got the suction point, ROS executed motion plan with MoveIt software. At last the ROS driven the real robot to execute the suction manipulation.

Table 3: The Results with Gaussian Filter

| Threshold | RGB | RGB-D | RGB-Points |
|---|---|---|---|
| 0.98 | 0.7853 | 0.8110 | 0.9574 |
| 0.85 | 0.5970 | 0.6717 | 0.7489 |

We used color images and point clouds as the input of U-net and executed some experiments in this platform. The hardware platform was a WidowX Robot Arm (Produced by Trossen Robotics, USA), which was shown in Figure 6. The results showed that the robot could suck and move all the cylinder objects in cluttered scenario efficiently.

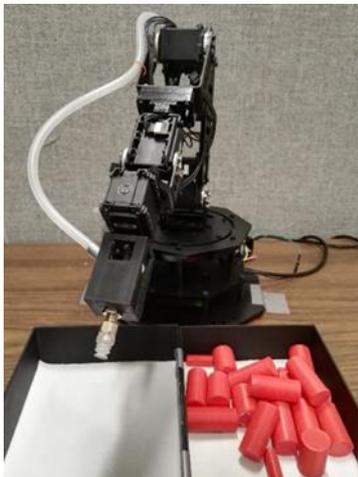

Figure 6: The Suction Grasp Platform

## V. CONCLUSION

This paper proposed a new method which used a region detection convolution framework to combine RGB and Points information to detect picking point in clutter. Experimental results demonstrated that U-net framework could also be used for picking point detection with a high efficiency as it revealed in image segmentation. This method predicted grasping region without recognition and pose estimation which was very appropriate in object-agnostic scenario. The results were also shown that combining color image and point clouds could get a high performance with the same structure of neural networks.

ACKNOWLEDGMENT

This work was supported by National Natural Science Foundation of China.